\documentclass[10pt,twocolumn,letterpaper]{article}
\usepackage{wacv}

\usepackage[utf8]{inputenc} % allow utf-8 input
\usepackage[T1]{fontenc}    % use 8-bit T1 fonts
\usepackage{url}            % simple URL typesetting
\usepackage{booktabs}       % professional-quality tables
\usepackage{amsfonts}       % blackboard math symbols
\usepackage{nicefrac}       % compact symbols for 1/2, etc.
\usepackage{microtype}      % microtypography
\usepackage{color, colortbl}
\usepackage{verbatim}
\usepackage{multirow}
\usepackage{amsmath,amssymb}
\usepackage{mathtools}
\usepackage{arydshln}
\usepackage[pdftex]{graphicx}
\usepackage[accsupp]{axessibility}
\usepackage{epsfig}
\usepackage{times}
\usepackage{paralist}
\usepackage{comment}
\usepackage[linesnumbered, ruled, vlined]{algorithm2e}
\usepackage{times}
\usepackage{epsfig}
\usepackage{relsize}
\usepackage{float}
\usepackage[noend]{algpseudocode}
\usepackage{lipsum}
\usepackage{widetext}
\usepackage{booktabs, makecell, multirow, tabularx}
\usepackage[table, usenames, dvipsnames]{xcolor}
\usepackage{authblk}

\usepackage[font=small, skip=0pt]{caption}
\usepackage[font=small, skip=0pt]{subcaption}

\usepackage[pagebackref=true,breaklinks=true,colorlinks,bookmarks=false]{hyperref}

\def\wacvPaperID{396} % Enter the WACV Paper ID here

\wacvfinalcopy % *** Uncomment this line for the final submission

\ifwacvfinal
\fi

\ifwacvfinal
\usepackage[breaklinks=true,bookmarks=false]{hyperref}
\else
\usepackage[pagebackref=true,breaklinks=true,colorlinks,bookmarks=false]{hyperref}
\fi

\begin{document}
%%%%%%%%% TITLE
%\title{\LaTeX\ Author Guidelines for WACV Proceedings}
\title{\textsc{HierMatch}: Leveraging Label Hierarchies for \\ Improving Semi-Supervised Learning}

\author{Ashima Garg \qquad Shaurya Bagga \qquad Yashvardhan Singh \qquad Saket Anand}
% \author{Bob Jones}
% \author{Alice Smith}\\
% \author{Bob Jones}
\affil{IIIT-Delhi, India \\ {\tt\small {\{ashimag, shaurya17104, 	
yashvardhan17123, anands\}}@iiitd.ac.in}}

%\affil{Affiliation}
\maketitle
\ifwacvfinal
\thispagestyle{empty}
\fi

%%%%%%%%% ABSTRACT
\begin{abstract}

Semi-supervised learning approaches have emerged as an active area of research to combat the challenge of obtaining large amounts of annotated data. Towards the goal of improving the performance of semi-supervised learning methods, we propose a novel framework, \textsc{HierMatch}, a semi-supervised approach that leverages hierarchical information to reduce labeling costs and performs as well as a vanilla semi-supervised learning method. Hierarchical information is often available as prior knowledge in the form of coarse labels (e.g., woodpeckers) for images with fine-grained labels (e.g., downy woodpeckers or golden-fronted woodpeckers).
However, the use of supervision using coarse-category labels to improve semi-supervised techniques has not been explored. In the absence of fine-grained labels, \textsc{HierMatch} exploits the label hierarchy and uses coarse class labels as a weak supervisory signal. Additionally, \textsc{HierMatch} is a generic-approach to improve any semi-supervised learning framework, we demonstrate this using our results on recent state-of-the-art techniques MixMatch and FixMatch. %Additionally, \textsc{HierMatch} builds on top of the popular semi-supervised learning framework MixMatch and learns a hierarchically-informed classifier. 
We evaluate the efficacy of \textsc{HierMatch} on two benchmark datasets, namely CIFAR-100 and NABirds. \textsc{HierMatch} can reduce the usage of fine-grained labels by 50\% on CIFAR-100 with only a marginal drop of 0.59\% in top-1 accuracy as compared to MixMatch. Code:  \url{https://github.com/07Agarg/HIERMATCH}
\end{abstract}
 %It is much easier to annotate a dataset at a coarse-level than it is to gather fine-grained annotations. 
%%%%%%%%% BODY TEXT
\section{Introduction}\label{sec:intro}
%Semi-supervised at hierarchical levels ... %
%{\color{red}\textbf{Need for supervision.}}
Recent achievements in deep learning can largely be attributed to the use of vast collections of labeled data. Data annotation is often tedious and time-consuming, and many applications such as fine-grained visual classification (FGVC) \cite{fu_2017_cvpr, zhuang_2020_AAAI, Hwang_2012_NIPS} and medical diagnostics require inputs from field experts for data annotation, thus increasing the cost of annotation. Semi-supervised learning (SSL) methods \cite{ sohn_2020_NeurIPS_fixmatch, Berthelot_2020_ICLR_ReMixMatch, berthelot_2019_NeurIPS, pi-model, mean-teacher} have emerged as a practical approach to overcome this costly requirement of supervised learning by leveraging vast quantities of unlabeled data along with limited supervision from labeled data. Acquiring large amounts of unlabeled data usually results in very little or no additional cost.
%%% Working figure
% \begin{figure}[tbp]
% \centering
% %\includepdf[page=1, scale=0.43]{Intro_diagram.pdf}
% %\centerline{\includegraphics[scale=0.5, trim=2.0cm 10cm 4.5cm 0cm]{intro_diagram.eps}}
% \includegraphics[width=0.48\textwidth]{Intro_diagram_cropped.pdf}
% \caption{A schematic illustration depicting the motivation of our idea. The upper network in the image shows the standard setting of semi-supervised methods. The algorithmic design of our proposed model (lower network) uses different levels of hierarchical labels to reduce the annotation cost of fine-grained labels while maintaining similar classification accuracy at the finest level. Labels change from coarse to fine with the increasing levels of hierarchy. 
% }
% \label{fig:fig1}
% \end{figure}

\begin{figure*}[tbp]
	\centerline{\includegraphics[scale=0.445]{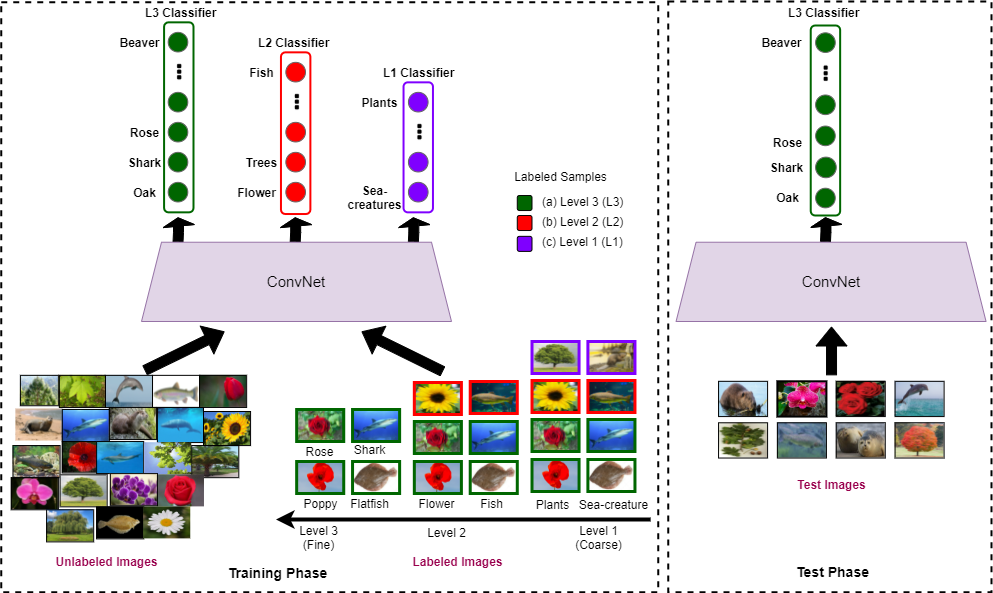}}
	\caption{\textbf{Overview of \textsc{HierMatch}}. Our proposed training strategy makes use of samples labeled at different hierarchical levels. Our network consists of different hierarchical classifiers (inspired by \cite{flamingo_2021_Chang}). At test time, we use only the finest-level classifier for classification. We assume that coarse-level labels are available for the samples labelled at a finer-level. \textsc{HierMatch} is a general-framework that can be used with any semi-supervised learning approach to incorporate information about label hierarchies. %We validate our claim by the results using MixMatch \cite{berthelot_2019_NeurIPS} and FixMatch \cite{sohn_2020_NeurIPS_fixmatch}.
	}
	\label{fig:proposed_approach}
\end{figure*}

% \begin{comment}
% \begin{figure*}[tbp]
% 	\centerline{\includegraphics[scale=0.45]{Proposed_approach.png}}
% 	\caption{\textbf{Overview of \textsc{HierMatch}}. Our proposed training strategy makes use of samples labeled at different hierarchical levels. Our network consists of different hierarchical classifiers (inspired from \cite{flamingo_2021_Chang}). At test time, we use only finest-level classifier for classification. We assume the coarse-level labels are available for the samples labelled at a finer-level. \textsc{HierMatch} is a general-framework that can be used with any semi-supervised learning approach to incorporate information about label hierarchies. %We validate our claim by the results using MixMatch \cite{berthelot_2019_NeurIPS} and FixMatch \cite{sohn_2020_NeurIPS_fixmatch}.
% 	}
% 	\label{fig:proposed_approach}
% \end{figure*}
% \end{comment}

% \begin{comment}
% {\color{red}\textbf{Existence of structure in data (hierarchy).}
% \begin{itemize}
%     \item Knowledge hierarchies / ontology reference (wordnet?). General way of organizing knowledge about data.  
%     \item Weaker supervision from coarser labels helps semi-supervised learning, i.e., with the labeled as well as the unlabeled data. How? Through consistency? By imposing other constraints or priors? 
%     \item Emphasize, with examples, that finer-grained labels are expensive and obtained by experts, but coarser grained labels can be cheaper and more readily available from lay-persons. 
% \end{itemize}
% }
% \end{comment}

Existing knowledge databases organize data by grouping related categories. WordNet \cite{miller_1998_wordnet}, a language dataset, categorizes the most generic object classes semantically in groups, which makes the hierarchical structure of labels easily accessible for most visual datasets. % Thus, given a finer-grained label, it is easy obtain coarser-level labels. 
While recent SSL methods %with consistency regularization and entropy minimization
\cite{sohn_2020_NeurIPS_fixmatch, berthelot_2019_NeurIPS} have made encouraging progress, the use of label hierarchies with SSL remains unexplored. As obtaining fine-grained labels typically needs expert knowledge, an effective way of reducing annotation cost is to make use of coarser-level supervision obtained from non-experts. % \textit{the use of coarse-level supervision to further reduce the annotations is not explored}. Semi-supervised learning approaches can benefit from coarse-level labels.  %that can act as a weaker form of supervision. %The coarse labels can assist SSL by imposing hierarchical priors on the structure of the feature space. 
% The hierarchical priors can be used to impose consistency, benefiting both labeled and unlabeled data. 
For instance, in NABirds \cite{van_2015_CVPR}, a fine-grained dataset, {\tt downy woodpeckers} and {\tt golden-fronted woodpeckers} are different fine-grained classes belonging to the same coarse class {\tt woodpecker}. A layperson can easily assign the label {\tt woodpecker} as compared to annotating it with fine-grained labels that can be done only by subject experts. We argue that by leveraging the readily available label hierarchies, we can trade-off expensive finer-grained labels annotated by experts in favor of cheaper, coarser-grained labels from non-experts, and yet devise semi-supervised techniques that perform nearly as well as baselines that use a larger number of fine-grained labels. 

Our claim builds upon the observation that all classes in a dataset do not differ equally from each other \cite{bertinetto_2020_CVPR}. By judiciously using label hierarchies, it is possible to train SSL-based classifiers with stronger priors despite weaker supervision (in the form of coarser labels) and achieve comparable fine-grained accuracy. For example, consider a dataset shown in Figure \ref{fig:proposed_approach} containing images of different kinds of {\tt Flower} and {\tt Fish} (Level 2 categories). The goal is to train a ConvNet based classifier for the Level 3 categories, i.e., \emph{types of} {\tt Flowers} and {\tt Fish}. By explicitly using label hierarchies during the SSL training, it should be possible to impose stronger priors and learn feature representations that will reduce confusion between unrelated Level 3 categories, e.g., between {\tt Sunflower} and {\tt Flatfish}. This observation is based on the argument that a human who is confident of an image being that of a {\tt Flower} (Level 2) is very unlikely to identify it as a {\tt Flatfish} (Level 3).

Existing techniques leverage the hierarchical priors on data in tasks such as object recognition \cite{marszalek_2007_CVPR, Hwang_2012_NIPS, grauman_2011_NIPS}, text-classification \cite{xu_2020_multimedia}, retrieval tasks \cite{barz_2019_WACV}, or by reducing the severity of mistakes \cite{bertinetto_2020_CVPR, karthik_2021_ICLR}. To our knowledge, ours is the first work that leverages hierarchical labels as knowledge priors in the semi-supervised setting leading to reduced labeling costs.

In this paper, we present an approach to incorporate label hierarchy in the SSL framework that trades off more expensive labels at the finest granularity of categories for cheaper labels at coarser granularity \emph{while maintaining classification performance} at the finest level.
% Our algorithm is a general approach to boost the performance of any semi-supervised learning method. Instead of naively applying any SSL approach, we apply the same algorithm on per-level hierarchical classifiers responsible for class predictions at respective hierarchical levels. In this paper, we maintain the performance of semi-supervised learning methods despite using fewer samples labeled at the finest-level. We reduce the samples labeled at the finest level and instead use an equal number of coarse labels. 
We refer to this training strategy as \textsc
{HierMatch} and test it in conjunction with two recently proposed semi-supervised learning techniques, MixMatch \cite{berthelot_2019_NeurIPS} and FixMatch \cite{sohn_2020_NeurIPS_fixmatch}. We demonstrate the effectiveness of \textsc{HierMatch} on datasets that offer two or more levels of granularity.

Our contributions are as follows:
\begin{itemize}
	\item We propose a novel framework, \textsc{HierMatch}, for improving the performance of a semi-supervised learning algorithm by exploiting the label hierarchies inherent in datasets.
	\item We experimentally validate that \textsc{HierMatch} achieves significant performance improvement on CIFAR-100 \cite{krizhevsky_2009_CIFAR100} and fine-grained NABirds \cite{van_2015_CVPR} dataset compared to MixMatch \cite{berthelot_2019_NeurIPS} by reducing 4000 and nearly 1000 fine-grained labels on CIFAR-100 and NABirds respectively. \textsc{HierMatch} can reduce the requirement to 50\% samples at the finest level with only 0.59\% performance drop on CIFAR-100. %Using FixMatch \cite{sohn_2020_NeurIPS_fixmatch} as SSL-algorithm, \textsc{HierMatch} is able to achieve nearly similar performance with 2k less fine-grained labels.
	\item We demonstrate the robustness of \textsc{HierMatch} to the choice of hierarchy with experiments on two different hierarchies for CIFAR-100: one with two levels, and the other with three. \textsc{HierMatch} improves upon baselines for both the settings. 
	\item We show that \textsc{HierMatch} does not need any additional hyperparameters. Furthermore, it works well with the hyperparameters as those set for the respective baseline SSL methods, therefore eliminating the need for any additional hyperparameter tuning.
\end{itemize}

The rest of the paper is organized as follows: %\begin{inparaenum}[i)]
Related work is discussed in Section \ref{sec:related_works}. In Section \ref{sec:approach}, we first discuss the preliminary ideas, followed by our proposed approach. Section \ref{sec:experiments} presents our experimental evaluations, followed by a discussion of results and their analysis in Section \ref{sec:analysis} and we finally conclude in Section \ref{sec:conc}.
%\end{inparaenum} 

\section{Related Work}\label{sec:related_works}
%\subsection{Fine-grained supervised methods}
\subsection{Semi-supervised learning methods}
Semi-supervised techniques utilize large amounts of unlabeled data by imposing a loss on predictions made on this unlabeled data. In the past, loss terms have been proposed to either nudge the model to make predictions with high confidence on unlabeled data (Entropy Minimization \cite{grandvalet_2005_NIPS}), or to predict a similar output distribution on perturbed inputs (consistency regularization \cite{pi-model, mean-teacher, berthelot_2019_NeurIPS, sohn_2020_NeurIPS_fixmatch}).

Laine et al. \cite{pi-model} incorporate consistency regularization loss in their \emph{$\pi$-Model}. For every image, labeled and unlabeled, they pass two augmentations of it through their network and force their network to make similar predictions on both (unsupervised loss term). For labeled images, they additionally impose cross-entropy loss (supervised loss term). As the training progresses and the model grows more confident in its predictions, they ramp up the weight of the unsupervised loss term. To reduce the number of forward passes per iteration, thus simplifying training, they also propose \emph{Temporal Ensembling} which alleviates the need for the second forward pass by maintaining an exponential moving average (EMA) of the model's previous predictions. However, \emph{Temporal Ensembling} introduces a large memory constraint that can pose challenges on large datasets. Subsequently,
Tarvainen et al. \cite{mean-teacher} improved upon the $\pi$-Model \cite{pi-model} by introducing a student-teacher model, wherein the weights of the teacher model are an EMA of the weights of the student model. They empirically show that using a teacher model improves the quality of targets produced that are used to enforce consistency regularization.
%Recent work by Berthelot et al \cite{berthelot_2019_NeurIPS} incorporates entropy minimization, consistency regularization and general regularization (input and model) in a single model, MixMatch. For every unlabeled image they produce K augmentations, which are then passed through the network to get K predictions. The K predictions are then averaged and then sharpened thus enforcing entropy minimization. Apart from data augmentations, MixMatch uses other regularization techniques such as MixUp \cite{mixup}. After processing the labeled and unlabeled datasets, the updated labeled data is used to compute the cross-entropy loss and the unlabeled data for computing the L2 loss (which enforces consistency regularization).\\

Zhai et al. \cite{s4l} bridge the gap between self-supervised learning and semi-supervised learning by proposing S4L. The basic S4L technique is simple, a self-supervised loss (eg. predicting rotations \cite{gidaris_2018_arxiv}) is computed on every image (labeled and unlabeled), and an additional cross-entropy loss is computed only on the labeled image.
%for every image (labeled or unlabeled), they compute a self-supervised loss (eg predicting rotations) and compute an additional cross-entropy loss on the labeled images. 
% They show their results on the ImageNet \cite{deng_2009_CVPR} dataset, where S4L outperforms, semi-supervised techniques such as Virtual Adversarial Training (VAT) \cite{vat}.

Berthelot et al. \cite{berthelot_2019_NeurIPS} combine different SSL techniques like consistency regularization, entropy minimization and general regularisation using mixup \cite{mixup} in their proposed framework, MixMatch. ReMixMatch \cite{Berthelot_2020_ICLR_ReMixMatch} by Berthelot et al. introduce improvements over MixMatch \cite{berthelot_2019_NeurIPS} by using distribution alignment and augmentation anchoring. They change consistency regularization by comparing weak augmentations of an image with its strong augmentation and show improvements. Sohn et al. \cite{sohn_2020_NeurIPS_fixmatch}, propose FixMatch, which combines different SSL components such as pseudo labeling and consistency regularization. Unlike ReMixMatch \cite{Berthelot_2020_ICLR_ReMixMatch} that uses \textit{sharpening} for consistency regularization, FixMatch \cite{sohn_2020_NeurIPS_fixmatch} uses \textit{pseudo-labels} of a weakly-augmented image against the strong augmentation of the same image. Despite its simplicity, FixMatch achieves state-of-the-art performance on most datasets and gives promising results in extremely label-scarce settings. %For a labeled image they apply a cross-entropy loss on a weakly augmented form of the image. For an unlabeled image, they predict a pseudo label on the weakly augmented image (using flip and shift augmentations) and retain the pseudo label if it passes a threshold. If the pseudo label is retained a cross-entropy loss is calculated on a strong augmentation (using RandAugment \cite{randaugment}, CTAugment \cite{Berthelot_2020_ICLR_ReMixMatch}) of the same image enforcing consistency regularization. Despite its simplicity, FixMatch achieves state-of-the-art performance on most datasets and gives promising results in extremely label-scarce settings. 
To the best of our knowledge, none of the current works in the domain of SSL leverage hierarchical information present in most datasets to boost their model performance. 

\subsection{Hierarchical classification methods}
The hierarchical structure present in data has been used in the supervised setting for various learning tasks, some of which we discuss in this section.  
% Here we discuss some techniques that exploit the hierarchical structure present in datasets for various tasks. 
%There have been works in the past that exploit the hierarchies present in the datasets to boost their model performance. 

%Devise \cite{frome_2013_NIPS} is a label-embedding method. It uses pretrained visual models to obtain the image embedding and a word2vec model, pretrained on Wikipedia, to obtain the label embedding. It attempts to learn feature and label embeddings that maximize the cosine similarity between them. %Barz and Denzler \cite{barz_2019_WACV} map samples onto a unit hypersphere and maximize the correlation based on the cosine similarity, where distance is encoded using the least common ancestor. 
Devise \cite{frome_2013_NIPS}, a label-embedding method, uses pretrained visual models to obtain the image embeddings and a pretrained word2vec model to obtain the corresponding label embeddings. They optimized a ranking loss such that the cosine similarity of the correct label should be higher than the similarity of any other label. Barz and Denzler \cite{barz_2019_WACV} map samples onto a unit hypersphere with distances based on the least common ancestor and maximize the correlation based on the cosine similarity. Bilal et al. \cite{bilal_2017_journal} analyze feature maps at different layers of the convnet block and errors in a simple classifier using the confusion matrix and branch out classification layers from the output of different convnet blocks in the decreasing order of coarseness at the last layer. 

Wu et al. \cite{wu_2016_ACM} use a multi-task loss function for each hierarchical level post the last fully connected layer. In the label inference step they smoothen the prediction values to enforce consistency across different semantic levels. Bertinetto et al. \cite{bertinetto_2020_CVPR} point out flaws with Top-$k$ error metrics that, for misclassified samples, all classes other than the ground-truth class are treated as equally wrong. To reduce the severity of the mistakes, they propose two novel loss functions, hierarchical cross-entropy (HXE) loss, and soft-labels loss. %Bertinetto et al. \cite{bertinetto_2020_CVPR} point out that the recent works in deep neural networks focus on performance metrics that treat all classes other than the ground truth class as equally wrong. Although this reduces the number of mistakes, it does not improve the quality of mistakes. They use the inherent hierarchies present in datasets to define the severity of a mistake as the height of the lowest common ancestor (LCA) between the misclassified prediction and the correct ground truth class in the hierarchy tree. They propose two novel loss functions, hierarchical cross-entropy (HXE) loss, and soft-labels loss, and empirically show their advantage over traditional cross-entropy loss in making a better quality of mistakes while having similar top-1 accuracy. Their soft-labels approach is also a label-embedding method that replaces the regular ground-truth one-hot encoded vector with soft-targets that encodes inter-class semantic information using the least common ancestor. 
Karthik et al. \cite{karthik_2021_ICLR} use a conditional risk minimization framework to correct mistakes by introducing hierarchical information at test time. Unlike conventional classifiers that maximize the probability of sample belonging to a class, they propose the use of a minimum risk class, defined based on the cost matrix.

%Wang et al. \cite{wang_2020_cvpr} explore the task of domain adaptation in fine-grained visual classification, wherein they exploit massive unlabeled data from target domain to improve performance on the source domain. They exploit the hierarchical structure inherent in datasets, integrate curriculum learning with adversarial learning to progressively improve domain adaptation from an easier setting of coarse-grained categories to a more difficult setting of fine-grained categories.

Wang et al. \cite{wang_2020_cvpr} exploit label hierarchies to tackle the task of domain adaptation in FGVC by integrating curriculum learning with adversarial learning to progressively improve domain adaptation in difficult setting of fine-grained categories.

\section{Proposed Approach}\label{sec:approach}

\subsection{Semi-Supervised Baselines \cite{berthelot_2019_NeurIPS, sohn_2020_NeurIPS_fixmatch}}
% \begin{comment}
% Common things in MixMatch and FixMatch - 
% \begin{itemize}
%     \item consistency regularisation
% \end{itemize}

% Differences in MixMatch and FixMatch - 
% \begin{itemize}
%     \item MixMatch integrates entropy minimisation, general regularisation (mixup)
%     \item FixMatch integrates pseudo labeling and thresholding
% \end{itemize}
% \end{comment}

Our technique leverages semi-supervised learning algorithms that incorporate consistency regularisation in their loss functions. We employ  MixMatch \cite{berthelot_2019_NeurIPS} and FixMatch \cite{sohn_2020_NeurIPS_fixmatch} as SSL techniques for our experiments. 
%\textbf{Intro to consistency regularisation} \\%

The semantic structure of an image is invariant to the augmentation applied.
Consequently, consistency regularisation encourages a model to make similar predictions on different augmentations of the same image. In the simplest formulation consistency is measured as the mean squared error between predictions on two augmentations of the same image. \cite{pi-model} introduce consistency loss, as an unsupervised loss, and compute this loss on both labeled as well as unlabeled images. The unsupervised loss is ramped up with time as the model grows more confident in its predictions.

MixMatch extends the $\pi$ model \cite{pi-model}  by proposing entropy minimisation via temperature scaling on unlabeled images. This forces the network to make more confident predictions for unlabeled images. MixMatch additionally employs MixUp \cite{mixup} as a regularizer to encourage convex prediction behaviour between convex combinations of samples. Akin to the $\pi$ model, MixMatch ramps up the unsupervised loss (consistency regularisation) as training progresses and the model becomes more confident in its predictions. MixMatch achieves much better results than the $\pi$ model \cite{pi-model}, albeit the number of hyperparameters and training time increase.
%\textbf{How fixmatch improves on the above techniques and other features of fixmatch, also mention different consistency regularisation}\\%

FixMatch \cite{sohn_2020_NeurIPS_fixmatch} improves on the $\pi$ model \cite{pi-model} by using pseudo labeling and changing the loss in consistency regularisation. FixMatch produces both a weak and a strong augmentation of every unlabeled image. The prediction of the network on the weakly augmented image is used as a pseudo label for the image if the predicted class probability is higher than some threshold value. This pseudo label is then used as the ground truth label for the prediction of the network on the strongly-augmented image. 

FixMatch enforces consistency regularisation only when the model is confident and does away with the need for time based consistency regularisation as is in $\pi$ model \cite{pi-model} and MixMatch \cite{berthelot_2019_NeurIPS}. FixMatch uses less hyperparameters than MixMatch, makes fewer forward passes through the network than MixMatch \cite{berthelot_2019_NeurIPS}, and has improved results by using pseudo-labels on unlabeled images, confidence thresholding, and introducing strong augmentations in consistency regularisation. %FixMatch has achieved state-of-the-art results in label scarce settings i.e. in CIFAR-10 they achieve an accuracy of 78\%, only by using 1 label per class.

\subsection{Hierarchical Classification \cite{flamingo_2021_Chang}} \label{flamingo-section}
Chang et al. \cite{flamingo_2021_Chang}, by means of a comprehensive human study, conclude that the label for an image is subjective depending on expertise in the domain and that most people prefer multi-granularity labels. They suggest that someone familiar with birds would assign the fine-grained label {\tt Flamingo} to an image containing a flamingo, however, a layperson would assign the coarse label {\tt Bird} to the very same image. An architecture consisting of $K$ independent label classifiers, one for each hierarchical level, forked from a common feature extraction backbone has been used before \cite{wang_2020_cvpr} towards the end of multi-granularity classification. The authors of \cite{flamingo_2021_Chang} establish that jointly optimizing for both coarse-grained and fine-grained classification equally on the aforementioned architecture degrades fine-grained classification performance while improving coarse-grained classification performance. Based on this finding, they propose disentangling of features fed into the independent label classifiers. They divide the feature vector $f$ equally into $K$ unique parts denoted as $f_k$. Each independent label classifier uses its features as well as features of classifiers at levels finer than itself to make predictions. Additionally, gradients from a label classifier are stopped from flowing to finer features during backpropagation and only flow to the features of this classifier. 
% \begin{comment}
% Thus, if ${\cal{G}}^{k}(.)$ is the classifier for $k$-th level hierarchy, then prediction from ${\cal{G}}^{k}(.)$ is given by: 
% \begin{equation} \label{eq:flamingo-classification}
% \hat{y}^k = {\cal{G}}^{k}(\text{CONCAT}(f_{k}, \Gamma(f_{k+1}), \ldots ,\Gamma(f_{K}))
% \end{equation}
% where $\Gamma(.)$ is used to represent the stop gradients and $K$ is the finest hierarchical level. 
% \end{comment}
This results in coarse grained classifiers enjoying the benefits of fine grained features while ensuring fine grained features remain unaffected by coarse grained supervision. 

\subsection{\textbf{\textsc{HierMatch}}}
% Given a training dataset with samples $\cal{D}$ = $\{ \mb{x_1} , \mb{x_2}, \mb{x_3}, \ldots \mb{x_n} \}$ in $\mathbb{R}^D$. 
Let $H$ be the number of hierarchical levels in the dataset and 
%$\{h^i\}_{i=1}^{H}$ be a set that represents the number of classes at different levels of hierarchy.
let $K^i$ denote the number of classes at the $i$-th level of the hierarchy, where $i=1$ is the coarsest and $i=H$ is the finest-level of the hierarchy. We focus on leveraging prior knowledge available in the hierarchical structure of data. Thus, for an image labeled at the hierarchical-level $i$ as $y^i$, we assume the image also has label information  available for all coarser $i-1$ levels as $\{y^{j}\}_{j=1}^{i-1}$.

Conventional semi-supervised learning frameworks use a set of unlabeled data and a set of labeled data. Labels for images in the labeled set are available only at the finest level of the hierarchy. In our setting of hierarchical semi-supervised learning, we have a set of labeled data for every level of the hierarchy and a set of unlabeled data.

Let ${\mathcal{X}}^{h}$ = $\{ (x_b, y_b^{h}) : b \in \mathcal{B}^h \}$ be the set of labeled data where $\mathcal{B}^h$ is the index set of all images that have labels at level $h\in [1,2,\ldots,H]$ and $y_b^h$ takes one of $K^h$ values. Each image $x_b$ was annotated at some level $j \geq h$  as $y_b^{j}$, which is its label at the $j$-th level of the hierarchy. The knowledge of labels $\{y_b^1, \ldots, y_b^h, \ldots, y_b^j\}$ for $x_b$ follows from our assumption about the knowledge of the hierarchical structure of labels. Thus $x_b$ will feature in all sets ${\mathcal{X}}^{1}, \ldots, {\mathcal{X}}^{j}$. 
%Consequently, the set of data annotated at the finest level has labels for all hierarchical levels and is  A contained in the set ${\mathcal{X}}^{H}$ = $\{ (x_b, y_b^{H} : b \in \mathcal{B}^H \}$. 
% Let ${\cal{X}}^{P}$ = $\{ (x_b^{P}, y_b^{P} : b \in (1, \ldots B) \}$ be a batch of data having labels for the $P$-th level of hierarchy. $x_b^{P}$ is the input sample and the $y_b^{P}$ is the corresponding ground truth label at the $i$-th level of hierarchy. Thus, $y_b^{P}$ takes one of $h^P$ values.  In one iteration, the batches of samples, having labels at different hierarchical levels, $\cal{X}^1,\dots,{X} ^H$ contribute to the supervised loss of our network.
Let the set of unlabeled data samples be $\cal{U}$ = $\{x_b : b \in \mathcal{B}^U \}$ where analogously $\mathcal{B}^U$ is the index set of \emph{unlabeled} images. These samples do not have label information at any hierarchical level. 

We denote the feature extractor by $\cal{F}(\cdot)$ and the label classifiers corresponding to different hierarchical levels by ${{\cal{G}}^1}(\cdot)$, ${{\cal{G}}^2}(\cdot)$, \ldots, ${{\cal{G}}^H}(\cdot)$, each of which fork from the output of $\cal{F}(\cdot)$. The labeled data for classifier ${\cal{G}}^{\ell}$ consists of all the samples that have a label at the $\ell$-th hierarchical level.
% \begin{comment}
% which is given below as the union of set of samples annotated at levels $h \ge \ell $.
% \begin{equation}
% {\mathcal{X}}^{\ell} = \bigcup_{h=\ell}^{H} \left({{\mathcal{X}}^h} = \left\{\left(x_b, \{y_b^{j}\}_{j=1}^h\right) : b \in \mathcal{B}^h\right\}\right)
% \end{equation}
% \end{comment}
% \begin{comment}
% \begin{equation}
% {\cal{L}}^{L_{i}} = \bigcup_{k=i}^{H}{{\cal{X}}^k} = \{(x_b^{i}, y_b^{i}) : b \in (1, \ldots, \sum_{j = i}^{H} n^{j})\}
% \end{equation}
% \begin{equation}
% {\cal{X}}^{L_{i}} = \{(x_b^{i}, y_b^{i}) : b \in (1, \ldots, \sum_{j = i}^{H} n^{j})\}
% \end{equation}
% \end{comment}
Unlabeled data for a classifier ${\cal{G}}^\ell$ comprises of all unlabeled samples $\mathcal{U}$ and samples labeled at levels coarser than $\ell$. This follows from the assumption that a sample labeled at a particular level $j$ does not have labels for any level finer than $j$. Thus, such a sample will be treated as an unlabeled sample by all the finer classifiers i.e. $\{{\cal{G}}^{j + 1},\dots,\mathcal{G}^{H}\}$. We collect these samples in a set defined as
\begin{equation}
{\cal{U}}^{\ell} = \left({\cal{X}}^{1} \setminus {\cal{X}}^{\ell}\right) \cup {\cal{U}} 
\end{equation}
%\begin{equation}
%{\cal{U}}^{L_{i}} = (\bigcup_{k=1}^{i-1}{{\cal{X}}^k}) \cup {\cal{U}} 
%\end{equation}

We build upon the hierarchical classification framework proposed by Chang et al. \cite{flamingo_2021_Chang} discussed in section \ref{flamingo-section}. Akin to their work, we disentangle the feature vectors used for predictions by the independent label classifiers. For an input image $x$, let the obtained feature vector $f = \mathcal{F}(x)$, which is divided into $H$ equal parts \{$f^1, \ldots, f^H$\}. Each vector $f^h$ corresponds to the $h$-th hierarchical level. Fine-grained features assist coarse-level classifiers in classification, however, during backpropagation gradients are stopped from flowing to the finer features to  avoid biasing them with coarse-level information. Thus, the final feature vector which the label classifier ${{\cal{G}}^h}(\cdot)$ uses for prediction is
\begin{equation}
f^h_{inp} = \text{CONCAT}(f^{h}, \Gamma(f^{h+1}), \ldots ,\Gamma(f^{H}))
\end{equation}
where $\Gamma(\cdot)$ represents stopping of gradients during backpropagation. 
%For level $i$, the feature vector that is used for the final classification is $f$ = $CONCAT(f_{i}, \Gamma(f_{i+1}), \ldots ,\Gamma(f_{H})$ where $\Gamma(.)$ represents the stop gradient. 
For each of the hierarchical classifiers, any semi-supervised learning algorithm can be applied. We use MixMatch \cite{berthelot_2019_NeurIPS} and FixMatch \cite{sohn_2020_NeurIPS_fixmatch} as examples of SSL algorithms to validate our claims. 
%For each of the hierarchical classifiers, we apply the MixMatch technique. 
We present the complete steps of \textsc{HierMatch} in Algorithm \ref{alg:our_algo}. 

\SetKwInput{KwInput}{Input}                % Set the Input
\SetKwInput{KwOutput}{Output}              % set the Output

%\begin{widetext}

\begin{algorithm*}
\caption{\textsc{HierMatch} Algorithm}
  \DontPrintSemicolon
  \label{alg:our_algo}
  \SetAlgoLined
  
  \KwInput {$\forall h \in \{1, \ldots, H\}  {\cal{X}}^{h}$ = $\{ (x_b^{h}, y_b^{h}) : b \in {\cal{B}}^h \}$ set of labeled data with labels till $h$-th level of hierarchies; \\
  $\cal{U}$ = $\{x_b : b \in {\cal{B}}^U\}$ set of unlabeled data points; Unsupervised loss weight $\lambda_u$; number of hierarchical levels in the data $H$. %sharpening temperature $\cal{T}$; number of augmentations $\cal{K}$; Beta distribution parameter $\alpha$;
  }
  \For{$\ell$=$\{1 \ldots H \}$}{
%    ${\cal{X}}^{\ell}$ = $\bigcup_{k=\ell}^{H}{{\cal{X}}^k}$\tcp*{Concatenate to form labeled data for level $\ell$}
    ${\cal{U}}^{\ell} = \left({\cal{X}}^{1} \setminus {\cal{X}}^{\ell}\right) \cup {\cal{U}} $\tcp*{Concatenate to form unlabeled data for level $\ell$}
  }
  \For{$e$=$\{1 \ldots MaxEpochs\}$}{
    \For{$t$=$\{1 \ldots MaxIters\}$}{
        ${\cal{L}}_{t}$ = $0$ \;
        %$\cal{L_{X^{L}}}$
        \For{$\ell$=$\{1 \ldots H\}$}{
        ${\cal{L}}_{{\cal{X}}^{\ell}_{t}}$, ${\cal{L}}_{{\cal{U}}^{\ell}_{t}}$ = \text{SSLAlgo} (${\cal{X}}^{\ell}_{t}$, ${\cal{U}}^{\ell}_{t}$) \label{algline:loss}\;
        \tcp*{Apply SSL algo. at level $\ell$ to calculate sup. \& unsup. loss}
        ${\cal{L}}_{t}$ += (${\cal{L}}_{{\cal{X}}^{\ell}_{t}}$ + $\lambda_u  {\cal{L}}_{{\cal{U}}^{\ell}_{t}}$) \tcp*{Calculate total loss}  
        }
        $\theta$ = \text{Optimize} (${\cal{L}}_{t}$, $\theta$) \tcp*{Update model parameters} 
    }
  }
%Use the trained model $\cal{{G}^{\text{$H$}}{(\cal{F(.)}})}$ for classification at the finest-level.
Use ${\cal{{G}^{\text{$H$}}}}(f^{H})$ for classification at the finest-level.
\end{algorithm*}

\section{Experiments}\label{sec:experiments}
\subsection{Datasets}
%\noindent 
We evaluate \textsc{HierMatch} on the following datasets:

\noindent \textbf{CIFAR-100} \cite{krizhevsky_2009_CIFAR100} - is a standard benchmark dataset which is used by many recent SSL techniques
\cite{Berthelot_2020_ICLR_ReMixMatch, berthelot_2019_NeurIPS, sohn_2020_NeurIPS_fixmatch}. It contains images of size $32\times 32$ belonging to one of 100 classes. Primarily, we use a two-level hierarchy, having 20 classes at level 1 and 100 classes at level 2 for our experiments. %\yash{Let's add where we got this hierarchy from.} 
To demonstrate the robustness of our method to different choices of hierarchy, we also report results on a three-level hierarchy \cite{Garnot_arxiv_2020}, an extension to the two-level hierarchy, with 8 classes at level 1, 20 classes at level 2, and 100 classes at level 3. 
%\shaurya{We could mention that the 3 level hierarchy is an extension to the 2 level hierarchy}

\noindent \textbf{North American Birds} \cite{van_2015_CVPR} - is a popular benchmark dataset commonly used to evaluate the performance of fine-grained visual classification methods. The NABirds dataset has a total of 555 classes. We perform experiments with two different hierarchies. The first is a two level hierarchy with 555 classes at level 2 and 404 classes at level 1. The second is an extension of the first with 555 classes at level 3, 404 classes at level 2, and 50 classes at level 1. To our knowledge, this paper is the first to evaluate semi-supervised learning methods on the large-scale fine-grained NABirds dataset. 

\subsection{Experimental Setup}
For both datasets, we split the training data into train and validation sets. We use 5000 samples of CIFAR-100 and 4972 samples of NABirds as validation sets. %These splits can be found in table \ref{tab:dataset_statistics}. %We use $~20\%$ of the entire training data as labeled data and the remaining samples as unlabeled data for both settings. 
For CIFAR-100, we stratify our split to maintain an equal number of samples per class in the labeled set. NABirds is a highly imbalanced dataset with number of samples in a class ranging from as low as 4 to a maximum of 60. We split the training set into labeled and unlabeled sets while ensuring every class has at least one sample in the labeled set. When distributing the labeled samples across hierarchies in NABirds, we ensure that there is at least one sample per class at the finest level. %We fine-tune our hyperparameters on the validation set and 
We report the final accuracy on the test set for the model with the best validation accuracy. %We do not use any part/bounding box annotations for training.

On the CIFAR-100 dataset, we show results using both MixMatch \cite{berthelot_2019_NeurIPS} and FixMatch \cite{sohn_2020_NeurIPS_fixmatch} as SSL algorithms in \textsc{HierMatch} framework and denote them as \textsc{HierMatch (M)} and \textsc{HierMatch (F)} respectively. We use the WideResNet-28-8 architecture \cite{zagoruyko_2016_BMVC} having 23.5 million parameters as our backbone network, following the state-of-the-art SSL methods \cite{berthelot_2019_NeurIPS, sohn_2020_NeurIPS_fixmatch}. % This is the same architecture used by state-of-the-art semi-supervised learning baseline methods \cite{berthelot_2019_NeurIPS, sohn_2020_NeurIPS_fixmatch}. 
%The backbone outputs a feature vector of size 512, which is then (nearly) equally divided between two or three levels of hierarchies to be fed into the hierarchical label classifiers. %A batch size of 64 is used. 

For a fair comparison, we use the same experimental settings on CIFAR-100 for \textsc{HierMatch (M)} and \textsc{HierMatch (F)} as is used for baseline MixMatch \cite{berthelot_2019_NeurIPS} and FixMatch \cite{sohn_2020_NeurIPS_fixmatch} respectively. We follow \cite{sohn_2020_NeurIPS_fixmatch} and use RandAugment \cite{cubuk_2020_randaugment} for data augmentation. 
%For both MixMatch \cite{berthelot_2019_NeurIPS} and FixMatch \cite{sohn_2020_NeurIPS_fixmatch} as SSL algorithm in \textsc{HierMatch} framework, for fair comparison, we keep the same experimental settings on CIFAR-100 as is for vanilla MixMatch \cite{berthelot_2019_NeurIPS} and vanilla FixMatch \cite{sohn_2020_NeurIPS_fixmatch} respectively. For FixMatch, we choose RandAugment \cite{cubuk_2020_randaugment}. %We use random pad and crop, followed by random flip as the data-augmentations techniques. We use the Adam optimizer \cite{Adam} with a fixed learning rate of 0.002 and $\beta$=(0.9, 0.99), with a weight decay of
%We use random pad and crop, followed by random flip as the data-augmentations techniques. We use the Adam optimizer \cite{Adam} with a fixed learning rate of 0.002 and $\beta$=(0.9, 0.99). A batch size of 64 is used. We set $\lambda_{u}$ to 150, same as \cite{berthelot_2019_NeurIPS}. The backbone outputs a feature vector of size 512, which is then equally divided between two (or three) levels of hierarchies to be fed into the hierarchical label classifiers. All the models are trained for 500 epochs with one epoch consisting of a total of 1024 iterations. 

For NABirds, we use the WideResNet-50-2 \cite{zagoruyko_2016_BMVC} (67.87 million parameters), pretrained on ImageNet \cite{ImageNet}, as our backbone network. We use only MixMatch \cite{berthelot_2019_NeurIPS} as semi-supervised technique in \textsc{HierMatch} framework. %\footnote{We do not report \textsc{HierMatch (F)} on NABirds due to computation time constraints. If accepted, we will include these results in the camera-ready version.}. %This network has a total of 67.87 million parameters. 
%We augment our data as is standard in the FGVC domain. The images are resized to 256x256 and randomly cropped to 224x224, followed by random horizontal flips and normalization. The backbone outputs a feature vector of size 2048, which is then divided into 682, 682 and 684 features across the 3 hierarchies (coarsest to finest). We use a batch size of 16. For the selection of the best set of hyper-parameters, we perform a sweep over $\lambda_{u}$, backbone learning rate, and classifier learning rate for 25 epochs over 20\% of the labeled set. Specifically we try $\lambda_{u}$ values $\in$ \{25, 75, 100, 150, 200\}, backbone learning rates $\in$ \{1e-4, 5e-4, 1e-5, 5e-5, 1e-6, 5e-6\} and classifier learning rates $\in$ \{1e-3, 1e-4, 5e-4, 1e-5, 5e-5, 1e-6\}. Our hyperparameter sweep found $\lambda_u$ of 100, backbone learning rate of 1e-5, and classifier learning rate of 1e-3 as the best set of hyperparameters. We use the Adam optimizer \cite{Adam} for both backbone and label classifiers. We use a batch size of 16. All models are trained for 250 epochs.
%We set the rest of the hyperparameters as $T$ = 0.5, $\alpha$ = 0.75. $K$ = 2. These remain the same for all the experiments on both NABirds and CIFAR-100. For all our MixMatch experiments, we linearly ramp up $\lambda_{u}$ to its maximum value over the entire training. 
We present further details of the complete set of hyperparameters used in the supplementary material. All our experiments were done using PyTorch \cite{pytorch}, and we used Weights \& Biases \cite{wandb} for experiment tracking, visualizations, and hyperparameter sweeps.

\subsection{Baselines}
We set out to show that exploiting hierarchies can reduce annotation costs in semi-supervised learning frameworks without compromising performance. We evaluate our approach against the following methods: \\ %\begin{inparaenum}[i)]
	%\begin{itemize}
	\textbf{Fully-supervised baseline} (\emph{Sup.-only})- 
	The backbone network with a single label classifier trained on all available labeled training samples. This setting represents the highest accuracy a model can achieve \textit{without} utilizing any hierarchical information in a purely supervised setting.\\
	\textbf{Fully-supervised using limited labeled samples
	%on 20\% of labeled samples 
	and no unlabeled samples} (\emph{Sup.-only}) - The backbone network with a single label classifier trained on a subset of the labeled dataset. This is the maximum performance one can achieve \textit{without} leveraging neither the hierarchical structure nor the unlabeled data.\\
	\textbf{Fully-supervised baseline using hierarchical network}
	
	\noindent (\emph{Hierarchical Sup.}) - we train the hierarchical network \cite{flamingo_2021_Chang} by utilizing the hierarchical structure inherent in the dataset, on the entire labeled data. This represents the highest achievable accuracy \textit{using the hierarchical coarse labels}.\\
	\textbf{MixMatch \cite{berthelot_2019_NeurIPS}}, \textbf{FixMatch \cite{sohn_2020_NeurIPS_fixmatch}} - The backbone network with a single label classifier trained using one of the SSL Algorithms MixMatch and FixMatch. In this setting, we use a subset of the entire available data as labeled, with labels at the finest-level only, and all of the unlabeled data.
% 	The backbone network with independent label classifiers per hierarchical level while utilizing the hierarchical structure inherent in the dataset, on the entire labeled data. This represents the highest achievable accuracy using the hierarchical coarse labels.
	%\textbf{MixMatch \cite{berthelot_2019_NeurIPS}} - The backbone network with a single label classifier trained using the semi-supervised algorithm MixMatch. In this setting, we use a subset of the entire available data as labeled, with labels at the finest-level only, as well as all of the unlabeled data.\\
	%\textbf{FixMatch \cite{sohn_2020_NeurIPS_fixmatch}} - The backbone network with a single label classifier trained using the semi-supervised algorithm FixMatch. In this setting, we use a subset of the entire available data as labeled, with labels at the finest-level only, as well as all of the unlabeled data.
%\end{itemize}
% \vspace{-4mm}
\subsection{Evaluation Protocols}
Unlike semi-supervised methods that use a fraction of data as a labeled set, our approach requires labels with different degrees of fineness. Therefore, we validate our approach by performing experiments with varying amounts of labeled data at different levels. We stick to other standard settings from SSL \cite{oliver_2018_NeurIPS, berthelot_2019_NeurIPS, Berthelot_2020_ICLR_ReMixMatch, sohn_2020_NeurIPS_fixmatch}. These include using a small validation set, using the same backbone network and keeping the same experimental settings across all the techniques per dataset.
As is typical in the semi-supervised learning setting, we report Top-1 and Top-5 accuracies on three-different folds of labeled data.

\section{Results and Analysis}\label{sec:analysis}

In the following section, an experiment is referred to using a tuple \texttt{($a,b,c$)}, where $a$ is the number (percentage) of labeled samples at level 3 (finest level), $b$ is the number (percentage) of samples labeled at level 2, and $c$ is the number (percentage) of samples labeled at level 1 (coarsest level). A sample labeled at a particular level $\ell$ contributes to the corpus of labeled samples at all levels coarser than the level $\ell$. Thus, for an experiment \texttt{($a,b,c$)}, we have a total of $a$ labeled samples at level 3, $a + b$ labeled samples at level 2, and $a + b + c$ labeled samples at level 1. A dash sign `-' in a tuple indicates the absence of that level of the hierarchy. For example, the tuple \texttt{($a,b,-$)} denotes a two level hierarchy with $a$ samples labeled at level $3$ and $b$ samples labeled only at level $2$. However, the tuple \texttt{($a,b,0$)} denotes a three level hierarchy with no additional samples labeled at level 1. In the absence of a hierarchy level $\ell^\prime$, the losses corresponding to that level (${\cal{L}}_{{\cal{X}}^{\ell^\prime}}$, ${\cal{L}}_{{\cal{U}}^{\ell^\prime}}$ in Line \ref{algline:loss} of Algo. \ref{alg:our_algo}) do not contribute at all.
% % \vspace{-5mm}

\subsection{Comparison with the baseline-methods}

Tables \ref{tab:CIFAR100_Baseline_experiments} and \ref{tab:NABirds_Baseline_experiments} present the comparisons of our proposed algorithm against the baselines. 
\textsc{HierMatch} outperforms the standard MixMatch algorithm on both datasets. Using the 2-level hierarchy for CIFAR-100, \textsc{HierMatch} increases top-1 accuracy  by 1.73\%  %and top-5 accuracy increases by 1.73\% 
while using the same number of samples labeled at the finest level. Using the 3-level hierarchy on NAbirds, the improvement in the top-1 accuracy is 1.02\%. %and the top-5 accuracy by 0.08\%. 

\noindent \textbf{Can \textsc{HierMatch} maintain performance, by leveraging hierarchical labels, despite reduction in number of samples labeled at the finest level?} \\
We use MixMatch and FixMatch with 10k labeled samples at the finest level as our baselines to answer this question. We reduce the number of samples labeled at the finest level and use an equal number of samples labeled at coarser levels instead. We report results on CIFAR-100 for \textsc{HierMatch (M)} in Table \ref{tab:CIFAR100_mixmatch_equal_ablations} and for \textsc{HierMatch (F)} in Table \ref{tab:CIFAR100_fixmatch_equal_ablations}. In the \texttt{(8k, 2k, -)} setting, using 2k less labeled samples at the finest level, \textsc{HierMatch (M)} improves upon the \texttt{(10k, -, -)} (i.e., the baseline MixMatch) in top-1 accuracy by 0.93\% and in top-5 accuracy by 0.3\%. In the \texttt{(5k, 5k, -)} setting, despite using 50\% less labeled samples at the finest level, our technique drops by only 0.59\%. In the \texttt{(7k, 2k, 1k)} setting, we obtain nearly similar performance as that of the baseline MixMatch. In domains where obtaining labels at the finest level is extremely expensive leveraging our technique can significantly reduce annotation costs whilst achieving similar performance. 
\begin{table}
\begin{center}
\small
%\resizebox{0.49\textwidth}{!}{
\begin{tabular}{l|c|cc} 
\hline
Methods                 & \begin{tabular}[c]{@{}c@{}}Labeled \\ Samples\end{tabular} & Top-1 (\%)        & Top-5 (\%)         \\
\cmidrule(l{2mm}r{2mm}){1-1} \cmidrule(l{2mm}r{2mm}){2-2} \cmidrule(l{2mm}r{2mm}){3-4}
Sup.-only              & (45k, -, -) & 79.98          & 93.91           \\
Sup.-only              & (10k, -, -) & 61.37          & 83.24           \\
Hierarchical Sup.      & (45k, 0, -) & 78.13          & 93.69           \\
\cmidrule(l{2mm}r{2mm}){1-1} \cmidrule(l{2mm}r{2mm}){2-2} \cmidrule(l{2mm}r{2mm}){3-4}
MixMatch               & (10k, -, -) & 71.69 $\pm$ 0.33          & 90.53   \\
\textsc{HierMatch (M)} & (10k, 0, -) & \textbf{73.42 $\pm$ 0.08} & \textbf{91.20 $\pm$ 0.10}  \\
FixMatch               & (10k, -, -) &  77.40 $\pm$ 0.12         & 94.3    \\
\textsc{HierMatch (F)} & (10k, 0, -) & \textbf{77.61 $\pm$ 0.06} & \textbf{94.70 $\pm$ 0.14}           \\
\hline
\end{tabular}
%}
\end{center}
\caption{Comparisons on the test sets of CIFAR-100 \cite{krizhevsky_2009_CIFAR100} with two-level hierarchy on the baseline methods. Top-1 and Top-5 accuracy are reported on the level 3 (finest) classifier. Approaches that do not use a hierarchical level are indicated with a dash mark `-'. %All the results are reported using our implementation.
}
\label{tab:CIFAR100_Baseline_experiments}
\end{table}
The results for NABirds are reported in Table \ref{tab:NABirds_equal_ablations}. We observe trends similar to those observed with CIFAR-100. In the settings \texttt{(15\%, 5\%, -)}, we reduce the number of samples at the finest level by 995 as compared to the baseline setting \texttt{(20\%, -, -)}, and improve on the top-1 and top-5 accuracy by 0.23\% and 0.59\% respectively. However, similar performance improvement is not observed in the much harder setting \texttt{(15\%, 0, 5\%)} since 5\% samples from level $3$ (finest level) are replaced by an equal number of samples at level $1$ (coarsest level). This can be attributed to the fact that we only have 50 categories at level $1$ which results in orders of magnitude less coverage of the input space. We observe high variations in the \textsc{HierMatch (M)} experiments of NABirds as total number of samples in fine-grained categories are in the range from 4 to 60. The encouraging performance of \textsc{HierMatch (M)} on NABirds, a challenging and a highly imbalanced dataset, speaks to the robustness of our algorithm. 
%\newcolumntype{C}{c<{\kern\tabcolsep}@{}}
%\rowcolor[gray]{0.9}[6pt][6pt]
%cyan!25!white
\begin{table}
\begin{center}
%\begin{tabular}{@{}ccc|cc@{}}
%\resizebox{0.49\textwidth}{!}{
\small
\begin{tabular}{c|cc} 
\hline
Labeled Samples & \begin{tabular}[c]{@{}c@{}}Top-1 (\%) \end{tabular} & \begin{tabular}[c]{@{}c@{}}Top-5 (\%)\end{tabular}          \\ \cmidrule(l{2mm}r{2mm}){1-1}				\cmidrule(l{2mm}r{2mm}){2-3}
 \rowcolor{lightgray} (10k, -, -)  & 71.69 $\pm$ 0.33                                                            & 90.53                                                                               \\
(10k, 0, -)  & 73.42 $\pm$ 0.08 & 91.20 $\pm$ 0.10 \\
(10k, 0, 0) & 72.85 $\pm$ 0.56 & 91.20 $\pm$ 0.20  \\ \cmidrule(l{2mm}r{2mm}){1-1}				\cmidrule(l{2mm}r{2mm}){2-3}
(8k, 2k, -) & 72.62 $\pm$ 0.10 & 90.83 $\pm$ 0.27 \\
(7k, 2k, 1k)  & 71.68 $\pm$ 0.59 & 90.41 $\pm$ 0.11 \\ \cmidrule(l{2mm}r{2mm}){1-1}				\cmidrule(l{2mm}r{2mm}){2-3}
(6k, 4k, -) & 71.93 $\pm$ 0.26 & 90.34 $\pm$ 0.09   \\
(6k, 2k, 2k)& 72.04 $\pm$ 0.37 & 90.33 $\pm$ 0.26 \\
\cmidrule(l{2mm}r{2mm}){1-1}				\cmidrule(l{2mm}r{2mm}){2-3}
(5k, 5k, -) & 71.10 $\pm$ 0.18 & 90.00 $\pm$ 0.09 \\
(5k, 3k, 2k) & 70.99 $\pm$ 0.24 & 89.73 $\pm$ 0.24 \\
\hline
\end{tabular}
%}
\end{center}
\caption{Performance comparison on CIFAR-100 \cite{krizhevsky_2009_CIFAR100} using MixMatch as SSL Algorithm. Accuracy on the level 3 (finest) classifier using different number of labeled samples from different hierarchies while keeping the total number of labeled samples as 10,000 across the two-levels of hierarchy. %Number of unlabeled samples as 35,000 remain the same in all the experiments. 
Approaches that do not use a hierarchical level are indicated with a dash mark `-'.} 
\label{tab:CIFAR100_mixmatch_equal_ablations}
\end{table}

\begin{table}
\begin{center}
%\begin{tabular}{@{}ccc|cc@{}} 
\small
\begin{tabular}{c|cc} 
\hline
\begin{tabular}[c]{@{}c@{}}Labeled \\ Samples\end{tabular} & \begin{tabular}[c]{@{}c@{}}Top-1 (\%) \end{tabular} & \begin{tabular}[c]{@{}c@{}}Top-5 (\%) \end{tabular}          \\ \cmidrule(l{2mm}r{2mm}){1-1}				\cmidrule(l{2mm}r{2mm}){2-3}
 \rowcolor{lightgray} (400, -, -) & 24.31 $\pm$ 0.22  & 37.87 $\pm$ 0.62  \\
(400, 0, -) & 31.39 $\pm$ 1.86  & 47.73 $\pm$ 1.50  \\
(300, 100, -) & 28.38 $\pm$ 0.65 & 45.32 $\pm$ 1.39  \\
\hline
\end{tabular}
\end{center}
\caption{Performance comparison on low-data CIFAR-100 \cite{krizhevsky_2009_CIFAR100} using MixMatch as SSL Algorithm. Accuracy on the level 2 (finest) classifier using different number of labeled samples from different hierarchies while keeping the total number of labeled samples as 400 across the two-levels of hierarchy. %Number of unlabeled samples as 35,000 remain the same in all the experiments. 
Approaches that do not use a hierarchical level are indicated with a dash mark `-'.} 
\label{tab:CIFAR100_small_equal_ablations}
\end{table}

\begin{table}
\begin{center}
%\begin{tabular}{@{}ccc|cc@{}} 
\small
\begin{tabular}{c|cc} 
\hline
\begin{tabular}[c]{@{}c@{}}Labeled\\ Samples\end{tabular} & \begin{tabular}[c]{@{}c@{}}Top-1 (\%)\end{tabular} & \begin{tabular}[c]{@{}c@{}}Top-5 (\%)\end{tabular}          \\ \cmidrule(l{2mm}r{2mm}){1-1}				\cmidrule(l{2mm}r{2mm}){2-3}
 \rowcolor{lightgray} (10k, -, -) & 77.40 $\pm$ 0.12 & 94.3   \\
(10k, 0, -)   &    77.61 $\pm$ 0.06 &  94.70 $\pm$ 0.14 \\
(8k, 2k, -) &  77.10 $\pm$ 0.16 & 94.53 $\pm$ 0.08 \\
(5k, 5k, -) &  76.06 $\pm$ 0.15 &  94.12 $\pm$ 0.12 \\
\hline
\end{tabular}
\end{center}
\caption{Performance comparison on CIFAR-100 \cite{krizhevsky_2009_CIFAR100} using FixMatch as SSL Algorithm. Accuracy on the level 3 (finest) classifier using different number of labeled samples from different hierarchies while keeping the total number of labeled samples as 10,000 across the two-levels of hierarchy. %Number of unlabeled samples as 35,000 remain the same in all the experiments. 
Approaches that do not use a hierarchical level are indicated with a dash mark `-'.} 
\label{tab:CIFAR100_fixmatch_equal_ablations}
\end{table}

\noindent \textbf{How effective is \textsc{\textsc{HierMatch}} if additional coarse-level labels are provided?}\\
We conduct a study on CIFAR-100 using MixMatch as SSL algorithm to establish the effectiveness of our algorithm with additional coarse-level labels, results of which are present in Table \ref{tab:CIFAR100_additional_ablations}. By keeping the number of samples constant at the finest level, we compare the results by varying the number of coarse-level samples. On all four settings, \textsc{HierMatch (M)} is able to achieve significant improvements when additional coarse-level labels are leveraged. This result confirms our claim that by incorporating additional hierarchical level information in an SSL framework with limited labels can boost its performance significantly, and obtaining such information is cost-effective.

\noindent \textbf{How does \textsc{\textsc{HierMatch}} perform in a small labeled data setting?}

% \vspace{-5 mm}
\noindent To test the performance of \textsc{HierMatch} in low-data setting, we use a total of 400 labeled samples from CIFAR-100 dataset i.e. 4 samples per finest-class are used. We record mean and standard deviation of label-scarce setting in Table \ref{tab:CIFAR100_small_equal_ablations}. Using baseline MixMatch \cite{berthelot_2019_NeurIPS} as SSL algorithm, we achieved an accuracy of 24.31\%. Using additional coarse-level labels of these 400 samples with \textsc{HierMatch} under the setting \texttt{(400, 0, -)} gives a significant boost of 7.08\% and 9.86\% in top-1 and top-5 accuracy metrics respectively. In the experiment setting \texttt{(300, 100, -)}, where we use only 3 samples per-fine grained class and an additional 100 coarse-labeled samples were used, we get an accuracy of 28.38\% which is nearly 4\% improvement over baseline MixMatch \cite{berthelot_2019_NeurIPS}. These improvements confirm that reducing fine-grained samples, and instead using coarser-samples can improve performance in low labeled data scenarios too. 
%Discuss results from Table \ref{tab:CIFAR100_small_equal_ablations}. 

\subsection{Effect of number of hierarchical levels}
In this subsection, we analyze the effect of varying the number of hierarchical levels for both CIFAR-100 and NABirds. We conduct experiments with hierarchical information from two and three hierarchical levels. The results of these are presented in Table \ref{tab:CIFAR100_mixmatch_equal_ablations}. For the experiments \texttt{(10k, -, -)}, \texttt{(10k, 0, -)} and \texttt{(10k, 0, 0)} where 10k samples remain the same at the finest-hierarchical level, the top-5 accuracy improves as we leverage more hierarchical information. We reduce 1000 finest-level samples from \texttt{(8k, 2k, -)}, increase them in the coarsest-level in \texttt{(7k, 2k, 1k)}, and observe a slight performance drop of 0.34\% in the top-1 accuracy.

The experiments \texttt{(6k, 4k, -)} and \texttt{(6k, 2k, 2k)} marginally differ by 0.11\% in top-1 and 0.01\% in top-5 accuracies respectively. The next set of experiments \texttt{(5k, 5k, -)} and \texttt{(5k, 3k, 2k)}, involving dropping of 2k samples from level 2 and increasing them in level 1, also differ marginally by 0.11\% and 0.27\% in top-1 and top-5 accuracies respectively. 
\begin{table}
\begin{center}
\small
\resizebox{0.49\textwidth}{!}{ 
\begin{tabular}{l|c|cc}
\hline
Methods                 & \begin{tabular}[c]{@{}c@{}}Labeled \\  Samples\end{tabular} & Top-1(\%)          & Top-5(\%)           \\
\cmidrule(l{2mm}r{2mm}){1-1} \cmidrule(l{2mm}r{2mm}){2-2} \cmidrule(l{2mm}r{2mm}){3-4}
Sup.-only              & (18957, -, -) & 69.94          & 88.37           \\
Sup.-only              & (4972 , -, -) & 55.71          & 50.86           \\
Hierarchical Sup. & (18957, 0, 0) & 71.39          & 88.78           \\
\cmidrule(l{2mm}r{2mm}){1-1} \cmidrule(l{2mm}r{2mm}){2-2} \cmidrule(l{2mm}r{2mm}){3-4}
MixMatch                & (4972, -, -) & 61.31 $\pm$ 0.23          & 81.86 $\pm$ 0.09           \\
\textsc{HierMatch (M)}     & (4972, 0, 0)  & \textbf{62.33 $\pm$ 0.24} 
& \textbf{82.20 $\pm$ 0.34}  
\\
\hline
\end{tabular}
}
\end{center}
\caption{Comparisons on the test sets of NABirds \cite{van_2015_CVPR} with three-level hierarchy on the baseline methods. Top-1 and Top-5 accuracy are reported on the fine-grained classifier. Approaches that do not use a hierarchical level are indicated with a dash mark `-'. %All the results are reported using our implementation.
}
\label{tab:NABirds_Baseline_experiments}
\end{table}

\begin{table}
\begin{center}
\small
%\resizebox{0.49\textwidth}{!}{
\begin{tabular}{c|cc} 
\hline
%\begin{tabular}[c]{@{}c@{}}Labeled \\ Samples\end{tabular} & 
Labeled Samples & 
\begin{tabular}[c]{@{}c@{}}Top-1 (\%)\end{tabular} & \begin{tabular}[c]{@{}c@{}}Top-5 (\%)\end{tabular}  \\  \cmidrule(l{2mm}r{2mm}){1-1} \cmidrule(l{2mm}r{2mm}){2-3}
\rowcolor{lightgray} (20\%, -, -)   & 61.31 $\pm$ 0.23  & 81.86 $\pm$ 0.09                                                                \\
%(4972 (20\%), 0, -) & 62.83 $\pm$ 0.48  & 83.36 $\pm$ 0.29 \\
%(4972 (20\%), 0, 0) & 62.33 $\pm$ 0.24  & 82.20 $\pm$ 0.34 \\
(20\%, 0, -) & 62.83 $\pm$ 0.48  & 83.36 $\pm$ 0.29 \\
(20\%, 0, 0) & 62.33 $\pm$ 0.24  & 82.20 $\pm$ 0.34 \\
\cmidrule(l{2mm}r{2mm}){1-1} \cmidrule(l{2mm}r{2mm}){2-3}
%(3977 (15\%), 995 (5\%), -) & 61.54 $\pm$ 0.22 & 82.45 $\pm$ 0.25 \\
%(3977 (15\%), 995 (5\%), 0) & 60.49 $\pm$ 0.81 & 80.03 $\pm$ 1.68 \\
%(3977 (15\%), 0, 995 (5\%)) & 59.33 $\pm$ 1.04 & 78.80 $\pm$ 1.82 \\
(15\%, 5\%, -) & 61.54 $\pm$ 0.22 & 82.45 $\pm$ 0.25 \\
(15\%, 5\%, 0) & 60.49 $\pm$ 0.81 & 80.03 $\pm$ 1.68 \\
(15\%, 0, 5\%) & 59.33 $\pm$ 1.04 & 78.80 $\pm$ 1.82 \\
\cmidrule(l{2mm}r{2mm}){1-1} \cmidrule(l{2mm}r{2mm}){2-3}
%(3475 (13\%), 1497 (7\%), -) &  60.03 $\pm$ 0.60 &  81.01 $\pm$ 0.89 \\
%(3475 (13\%), 991 (4\%), 506 (3\%)) & 59.39 $\pm$ 0.82 & 79.03 $\pm$ 1.50 \\
(13\%, 7\%, -) &  60.03 $\pm$ 0.60 &  81.01 $\pm$ 0.89 \\
(13\%, 4\%, 3\%) & 59.39 $\pm$ 0.82 & 79.03 $\pm$ 1.50 \\
\hline
\end{tabular}
%}
\end{center}
\caption{Performance comparison on NABirds \cite{van_2015_CVPR}. Top-1 and Top-5 Accuracy (\%) are reported on the level 3 classifier using different number of labeled samples from different hierarchies while keeping the total number of labeled samples as 20\% across the three-levels of hierarchy. %\% of samples in each level indicate number of labeled samples used from every class in that level. 
\% indicate the respective number of samples used 20\% - 4972, 15\% - 3977, 13\% - 3475, 7\% - 1497, 4\% - 991, and 3\% - 506.
%Number of unlabeled samples 18,957 used are same across all the experiments. 
Approaches that do not use a hierarchical level are indicated with a dash mark `-'.}
\label{tab:NABirds_equal_ablations}
\end{table}

\begin{table}
\small
\begin{center}
%\resizebox{0.48\textwidth}{!}{
\begin{tabular}{cc|cc}
\hline
\begin{tabular}[c]{@{}c@{}}Labeled \\ Samples\end{tabular} & \begin{tabular}[c]{@{}c@{}}Top-1 (\%) \end{tabular} & \begin{tabular}[c]{@{}c@{}}Labeled \\ Samples\end{tabular} & \begin{tabular}[c]{@{}c@{}}Top-1 (\%)\end{tabular}       \\ 
\cmidrule(l{2mm}r{2mm}){1-2} \cmidrule(l{2mm}r{2mm}){3-4}
(5k, 0, -)  &    67.92                 & (5k, 5k, -) &  71.21 \\
(6k, 0, -)  &    69.65                  & (6k, 4k, -)  & 71.72  \\
(8k, 0, -)  &   70.78                 & (8k, 2k, -)   & 72.61  \\
(10k, 0, -) &   73.51               & (10k, 5k, -)  &  74.91 \\
\hline
\end{tabular}
%}
\end{center}
\caption{Ablation study on CIFAR-100 \cite{krizhevsky_2009_CIFAR100} to show that the additional coarse-class level labels can indeed help improve the performance. Samples from fine-grained class (level 3) are kept constant while an increase in the additional number of coarse-grained (level 2) samples improve the performance. Accuracy is reported on the fine-grained classifier.}
\label{tab:CIFAR100_additional_ablations}
\end{table}

\section{Conclusion}\label{sec:conc}
We introduced \textsc{HierMatch}, a technique that leverages hierarchical structure often available with real-world datasets to empower existing semi-supervised learning methods. 
% We introduced \textsc{HierMatch}, a technique that combines ideas from semi-supervised learning with the important observation that most real-world datasets have some hierarchical structure to them. 
Through experiments for two different choices of hierarchies on both CIFAR-100 and NABirds we found that with fewer number of fine-grained labeled samples, \textsc{HierMatch} was able to achieve competitive or better performance than the baseline MixMatch technique. % despite substituting up to 50\% of samples labeled at the finest level with an equal number of samples labeled at coarser levels. 
This work establishes that using hierarchical priors boosts the performance of state-of-the-art semi-supervised learning methods on large datasets. For future work, we are interested in exploring other ideas from semi-supervised learning that might be more suitable for hierarchical structure of data as well as validating our technique on more complex datasets.

\section*{Acknowledgement}
This work was supported partly by SERB, Govt. of India, under grant no. CRG/2020/006049 and partly by the Infosys Center for Artificial Intelligence at IIIT-Delhi. 

\section*{Supplementary}\label{sec:supplementary}
% \documentclass[10pt, twocolumn, letterpaper]{article}
% %\usepackage[demo]{graphicx} # do not use this. 
% \usepackage{wacv}
% \usepackage{times}
% \usepackage{epsfig}
% \usepackage{graphicx}
% \usepackage{amsmath}
% \usepackage{amssymb}
% \usepackage{paralist}
% \usepackage{comment}
% %\usepackage{caption}
% \usepackage{subcaption}

% \def\wacvPaperID{396} % Enter the WACV Paper ID here

% %(2)
% \wacvfinalcopy % *** Uncomment this line for the final submission

% %(3)
% % Pages are numbered in submission mode, and unnumbered in camera-ready
% \ifwacvfinal
% \pagestyle{empty}
% \fi

% %%%%%%%%%%%%%%%%%%%%%%%%%%%%%%%%%%%%%%%%%%%%%%%%%%%%%%%%%%%%%%%%%%%%%%%%%%%%%%%%
% % If you comment hyperref and then uncomment it, you should delete
% % egpaper.aux before re-running latex.  (Or just hit 'q' on the first latex
% % run, let it finish, and you should be clear).
% \ifwacvfinal
% \usepackage[breaklinks=true,bookmarks=false]{hyperref}
% \else
% \usepackage[pagebackref=true,breaklinks=true,colorlinks,bookmarks=false]{hyperref}
% \fi

% \def\httilde{\mbox{\tt\raisebox{-.5ex}{\symbol{126}}}}

% \ifwacvfinal
% \thispagestyle{empty}
% \fi
% \begin{document}

\section{Hyperparameters}

\subsection{CIFAR-100 \cite{krizhevsky_2009_CIFAR100}}
\noindent Wideresnet-28-8 \cite{zagoruyko_2016_BMVC} architecture outputs a feature vector of size 512, which is equally divided between two-level hierarchy and nearly equal as 170, 170 and 172 (coarsest to finest) for three-level of hierarchies to be fed into the hierarchical label classifiers. A batch size of 64 is used. 
\\

\noindent \textbf{\textsc{HierMatch (M)}: } \\
We use random pad and crop, followed by random flip as the data-augmentations techniques. We use Adam optimizer \cite{Adam} with a fixed learning rate of 0.002 and $\beta$=(0.9, 0.99), with a weight decay of 4e-5. All the models are trained for 500 epochs with one epoch consisting of a total of 1024 iterations. We set unlabeled loss weight $\lambda_{u}$ to 150, same as \cite{berthelot_2019_NeurIPS}. \\

\noindent \textbf{\textsc{HierMatch (F)}: } \\
We use random horizontal flip followed by random crop for ``weak'' augmentation and additionally use RandAugment \cite{cubuk_2020_randaugment} for 
``strong'' augmentation of the image, following baseline FixMatch \cite{sohn_2020_NeurIPS_fixmatch} across all the experiments. We utilize SGD optimizer with a nesterov momentum of 0.9. Cosine learning rate scheduler with an initial learning rate of 0.03 is used. Unlabeled loss weight $\lambda_{u}$ is set to 1 throughout the training. We use unlabeled data ratio of 7 and set confidence threshold as 0.95.

\subsection{North-American Birds (NABirds) \cite{van_2015_CVPR}}
%This network has a total of 67.87 million parameters. 
\noindent \textbf{\textsc{HierMatch (M)}: }\\
\noindent We augment our data as is standard in the FGVC domain. The images are resized to 256x256 and randomly cropped to 224x224, followed by random horizontal flips and normalization. The backbone outputs a feature vector of size 2048, which is then divided into 682, 682 and 684 features across the 3 hierarchies (coarsest to finest). We use a batch size of 16. For the selection of the best set of hyper-parameters, we perform a sweep over $\lambda_{u}$, backbone learning rate, and classifier learning rate for 25 epochs over 20\% of the labeled set. Specifically we try $\lambda_{u}$ values $\in$ \{25, 75, 100, 150, 200\}, backbone learning rates $\in$ \{1e-4, 5e-4, 1e-5, 5e-5, 1e-6, 5e-6\} and classifier learning rates $\in$ \{1e-3, 1e-4, 5e-4, 1e-5, 5e-5, 1e-6\}. Our hyperparameter sweep found $\lambda_u$ of 100, backbone learning rate of 1e-5, and classifier learning rate of 1e-3 as the best set of hyperparameters. We use Adam optimizer \cite{Adam} for both backbone and label classifiers. We use a batch size of 16. All the models are trained for 250 epochs.\\ 
\noindent We set the rest of the MixMatch hyperparameters - temperature sharpening $T$ = 0.5, MixUp beta distribution parameter $\alpha$ = 0.75, and number of augmentations $K$ = 2. These remain the same for all the experiments on both NABirds and CIFAR-100. For all our MixMatch experiments, we linearly ramp up $\lambda_{u}$ to its maximum value over the entire training. 

% \begin{figure}[t]
%   \centering
%         \begin{subfigure}[!t]{0.50\linewidth}
%       \centerline{\includegraphics[width=\linewidth]{cifar100_max_probs_histogram.pdf}}
%           \caption{$n = 50, \eta = 1, \epsilon = 0$}
%       \end{subfigure}
       
%       \begin{subfigure}[!t]{0.50\linewidth}
%         \centerline{\includegraphics[width=\linewidth]{nabirds_max_probs_histogram.pdf}}
%         \caption{$n = 1000, \eta = 1, \epsilon = 0$}
%       \end{subfigure}

%   \caption{\small{Plots: (a) Loss vs Epochs, (b) Accuracy vs Epochs for both training and testing sets}}
%   \label{fig:1}
%   \vspace{-0.75em}
% \end{figure}

% \\ \\
\begin{figure}
\begin{minipage}[b]{.5\linewidth}
\centering
\includegraphics[width=\textwidth]{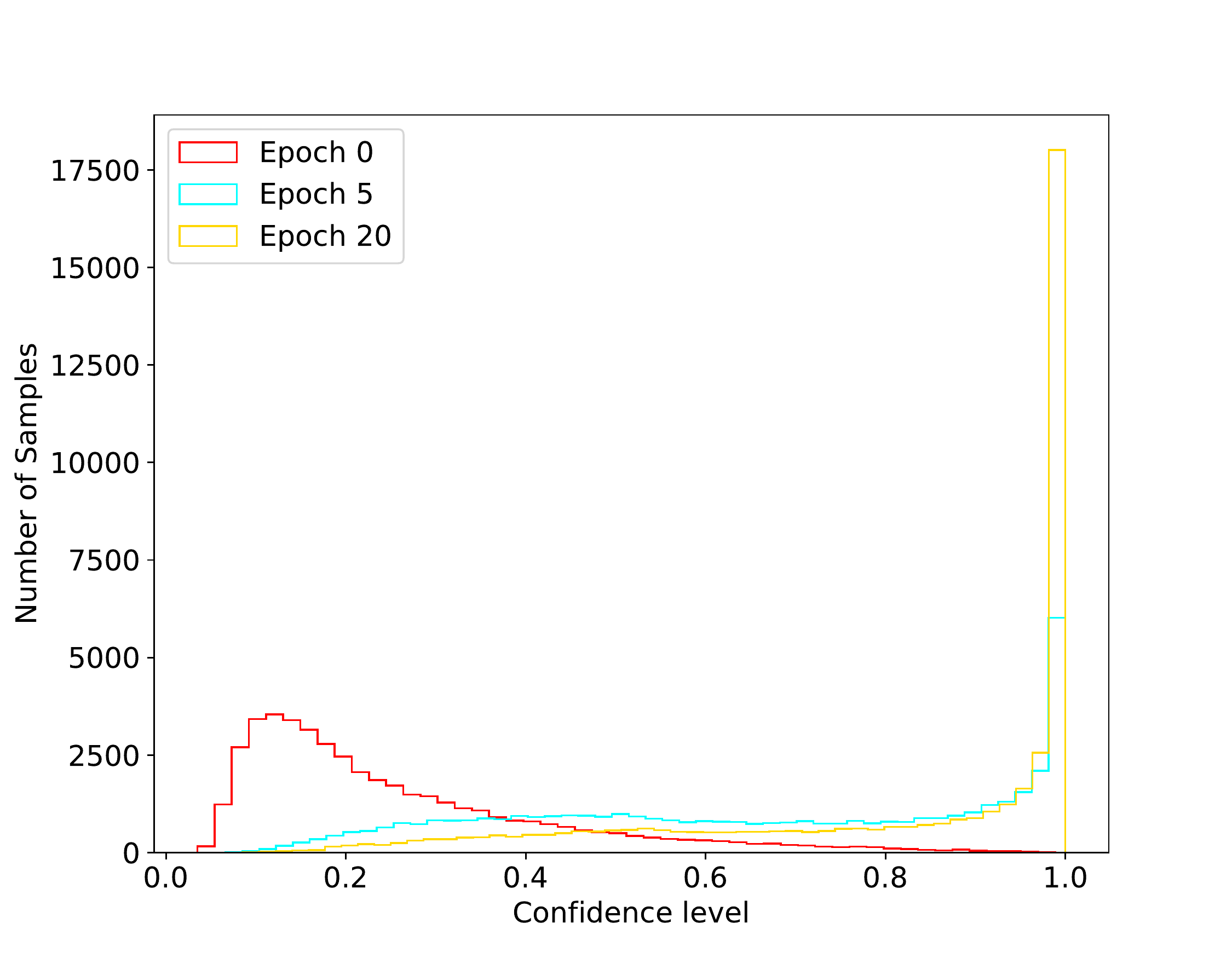}
\subcaption{CIFAR-100}\label{fig:cifar100}
\end{minipage}%
\begin{minipage}[b]{.5\linewidth}
\centering
\includegraphics[width=\textwidth]{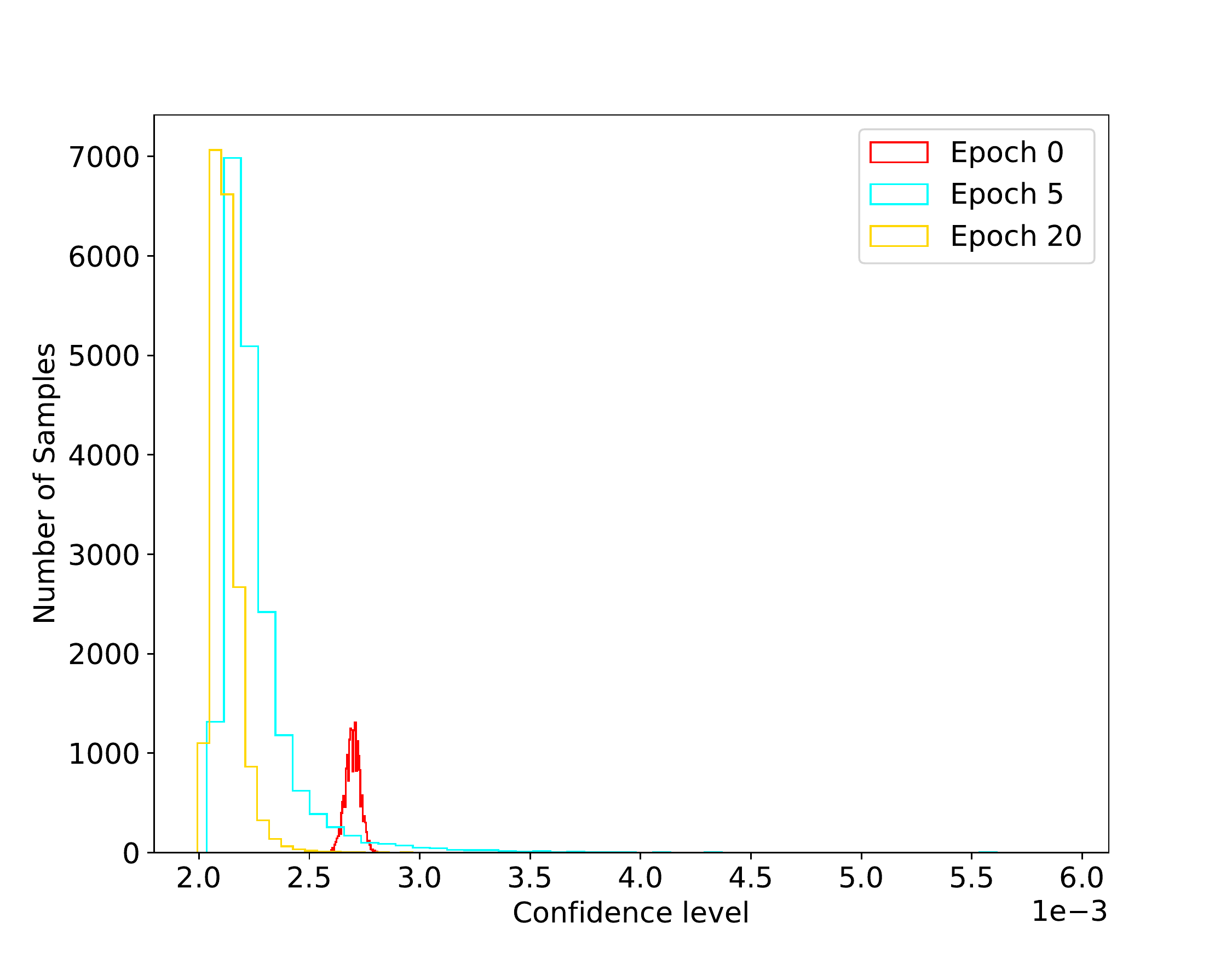}
\subcaption{NABirds}\label{fig:nabirds}
\end{minipage}
\caption{Histogram of confidence values on pseudo-labels of unlabeled data at different epochs on CIFAR-100 (top) and NABirds (bottom) dataset.}\label{fig:confidence_over_epochs}
\vspace{-2em}
\end{figure}

% \begin{figure}
%      \centering
%      \begin{subfigure}[b]{0.4\textwidth}
%          \centering
%          \includegraphics[width=\textwidth]{cifar100_max_probs_histogram.pdf}
%          \caption{CIFAR-100}
%          \label{fig:cifar100}
%      \end{subfigure}
%      \hfill
%      \begin{subfigure}[b]{0.4\textwidth}
%          \centering
%          \includegraphics[width=\textwidth]{nabirds_max_probs_histogram.pdf}
%          \caption{NABirds}
%          \label{fig:nabirds}
%      \end{subfigure}
%      \hfill
%     \caption{Histogram of confidence values on pseudo-labels of unlabeled data at different epochs on CIFAR-100 (top) and NABirds (bottom) dataset.}
%     \label{fig:confidence_over_epochs}
% \end{figure}
% \begin{figure}
%      \centering
%      \begin{subfigure}[b]{0.4\textwidth}
%          \centering
%          \includegraphics[width=\textwidth]{cifar100_max_probs_histogram.pdf}
%          \caption{CIFAR-100}
%          \label{fig:cifar100}
%      \end{subfigure}
%      \hfill
%      \begin{subfigure}[b]{0.4\textwidth}
%          \centering
%          \includegraphics[width=\textwidth]{nabirds_max_probs_histogram.pdf}
%          \caption{NABirds}
%          \label{fig:nabirds}
%      \end{subfigure}
%      \hfill
%     \caption{Histogram of confidence values on pseudo-labels of unlabeled data at different epochs on CIFAR-100 (top) and NABirds (bottom) dataset.}
%     \label{fig:confidence_over_epochs}
% \end{figure}

\noindent \textbf{\textsc{HierMatch (F)}: }\\
For labeled data and ``weak'' augmentation of an image, we use the same augmentations as is used for \textsc{HierMatch (M)} for NABirds. For ``strong'' augmentation, we additionally employ RandAugment similar to CIFAR-100. We use backbone WideResNet-50-2 \cite{zagoruyko_2016_BMVC} backbone, pretrained on ImageNet, as our backbone network. We sweep over backbone learning rates $\in$ \{1e-4, 5e-4, 5e-5, 1e-5]\}, classifier learning rates $\in$ \{1e-3, 1e-4, 5e-4, 1e-5, 5e-5, 1e-6\}, and threshold values $\in$ \{0.5, 0.7, 0.95\}. SGD optimizer with nestorov momentum of 0.9 was used. \\

\noindent FixMatch enforces consistency regularisation only when the model is confident. Interestingly, in our experiments of NABirds, we observe that with the best hyperparameters (backbone learning rate of 1e-5 and classifier learning rate of 1e-3), and with a minimum confidence threshold of 0.5, all of the most confident unlabeled samples fall below this threshold. Consequently,  the consistency regularization is not imposed at all, implying that the unlabeled loss is zero throughout the training phase while the training accuracy on labeled set overfits the dataset with 100\%. The resulting validation accuracy is similar to that of the Fully-supervised setting using limited labeled samples on NABirds. In Figure \ref{fig:cifar100} of CIFAR-100, as the training progresses the confidence values on unlabeled data improve whereas in Figure \ref{fig:nabirds} the confidence values on unlabeled data is too low and therefore, unlabeled data is not used at all while FixMatch training. %Also, FixMatch is sensitive to the choice of hyperparameters and thus, 
FixMatch requires both ``weak'' and ``strong'' augmentations using RandAugment \cite{cubuk_2020_randaugment} which requires more careful experimentation to design such augmentations for fine-grained datasets like NABirds. %More careful experimentation is needed to select the best choice of hyperparameters for FixMatch. 
Despite our hyperparameter sweeps, the baseline FixMatch \cite{sohn_2020_NeurIPS_fixmatch} gives poor performance on NABirds \cite{van_2015_CVPR}, so we do not report experimental evaluation of NABirds on FixMatch and \textsc{HierMatch (F)}.

{\small
\bibliographystyle{ieee_fullname}
\bibliography{bibliography}
}

\end{document}